# Can Large Language Models Reliably Code Qualitative Humanitarian Data? A Benchmark Study Against Human Expert Adjudication

Jerome Marston[1], Tino Kreutzer[1 2], Salomé Garnier[1], Ella Boone[2], Phuong N Pham[3], Patrick Vinck[3]

[1]Kobo Inc, [2] McGill University, [3] Harvard University

*This article is a preprint and has not been peer reviewed. The findings and conclusions should therefore be interpreted as preliminary.*

## Abstract

Data from affected populations are crucial for informing humanitarian response, but their value depends on timely and consistent interpretation of nuanced accounts of need. Humanitarian organizations often lack the staff, time, and specialist expertise required to analyze this information at scale. Large language models (LLMs) may expand this capacity, but their reliability for coding qualitative humanitarian data has not been directly established. This benchmark study compares 46 LLMs to a human Gold Standard using 150 high-fidelity synthetic humanitarian transcripts. Evaluation combined inter-rater reliability testing with Krippendorff's alpha, discrepancy analysis distinguishing correct, near-correct, and incorrect codes, and qualitative assessment across humanitarian-specific criteria including discrimination, complex needs hierarchies, and non-standard communication styles. The authors find that multiple LLMs can perform deductive coding at reliability levels comparable to experienced human coders, especially when structured prompts and reasoning-enabled configurations are used. At the same time, aggregate reliability metrics alone are insufficient for deployment decisions. Models varied in recognizing needs expressed indirectly, needs outside predefined categories, and protection-relevant concerns such as physical safety and discrimination. These findings suggest that LLMs can materially expand humanitarian analytical capacity, but not as substitutes for human judgment. Appropriate use requires structured codebooks, reasoning-enabled models, attention to theme-specific performance, and tiered oversight focused on categories where miscoding would have the greatest programmatic consequences. For sensitive humanitarian data, open-weights models deployed on self-hosted infrastructure may offer a viable path for combining analytical scalability with stronger data governance.

**Corresponding author:** Tino Kreutzer, tino.kreutzer@kobo.ngo

## Introduction

Qualitative data from crisis-affected populations, including interview transcripts, narrative survey responses, and community feedback records, carry crucial information that about the specific needs and perspectives of diverse individuals across varied situations. In humanitarian response, these nuanced data inform needs assessments, aid delivery, program prioritization, and accountability to affected communities. Systematic thematic coding, the interpretive process of assigning categorical labels to segments of text, is central to qualitative analysis but resource-intensive, requiring trained personnel, shared codebooks, and iterative quality control. In organizations operating under the constraints typical of humanitarian response, qualitative data are routinely collected in volumes that outpace the staff's analytical capacity, limiting the operational influence of this key information source.

Large language models (LLMs) offer a potential means of reducing this burden. Their capacity for contextual text classification has produced a growing body of research on LLM-assisted qualitative coding, with some studies reporting agreement with human coders approaching expert-level inter-rater reliability under structured conditions. However, this evidence has been generated almost entirely in non-humanitarian settings, including medical education, software engineering, and learning design, and its applicability to humanitarian data analysis, with its distinct linguistic diversity, thematic sensitivity, and accountability requirements, has not been directly established. The use of automated coding tools in humanitarian settings carries practical and ethical implications beyond technical performance (Kreutzer et al., 2025). Where coding outputs affect decisions about aid delivery, resource allocation or program response, errors carry costs borne disproportionately by already-vulnerable populations, and accountability for those errors cannot be delegated to automated systems.

This paper presents a systematic evaluation of 46 LLMs for qualitative humanitarian data coding, tested against a validated human Gold Standard across 48,300 coding iterations. Evaluation proceeds in three stages: inter-rater reliability testing using Krippendorff's alpha; a discrepancy analysis that distinguishes correct, near-correct, and entirely incorrect codes; and a qualitative assessment of model performance across criteria specific to humanitarian contexts, including recognition of discrimination, complex needs hierarchies, and non-standard communication styles. A dataset of 150 synthetic transcripts, designed to reflect the complexity and diversity of real humanitarian interview data, served as the test dataset, enabling rigorous and repeatable evaluation without exposing sensitive participant data to external systems.

The authors make four contributions: (1) the first direct evaluation of LLM coding reliability on humanitarian-context data, (2) a multi-stage evaluation framework that extends conventional inter-coder reliability testing to include qualitative error classification, (3) documentation of performance variation between LLMs by humanitarian theme and reasoning effort, and (4) evidence relevant to the feasibility of deploying open-weights models as privacy-respecting alternatives to proprietary systems for humanitarian organizations.

## Background and Literature Review

The use of LLMs in qualitative coding engages several methodological debates: what qualitative analysis requires interpretively, what automated systems can and cannot capture in textual meaning, and what accountability mechanisms are necessary when analysis outputs inform high-stakes decisions. This review synthesizes evidence on LLM performance in qualitative coding, examining reliability, the conditions under which performance is achieved, the limits of agreement-based evaluation, and the absence of direct evidence in humanitarian applications.

### Qualitative Thematic Coding: Interpretive Demands and Operational Constraints

Thematic coding is a core procedure in qualitative research, applied to interview transcripts, focus group records, and open-ended survey responses to identify, organize, and report patterns of meaning in text (Braun and Clarke, 2006). The process involves interpretive judgment rather than mechanical classification: coders must determine which themes are primary in a given text, how context modifies meaning, and how to handle cases that resist clear categorical assignment. Coding can be deductive, applying a predetermined codebook developed from theory or operational requirements, or inductive, with categories emerging through close reading of the data. Both approaches require more than surface-level text matching; the coder must assess not only what topics are present but what they mean in context and in relation to other content. Inter-coder reliability coefficients, among which Krippendorff's alpha and Cohen's kappa are the most widely used, provide standardized measures of agreement adjusted for chance, establishing whether independent coders apply categories consistently (Krippendorff, 2019; Gwet, 2014; Marzi et al., 2024). These metrics are necessary but not sufficient for coding quality: consistent agreement among coders who share the same interpretive limitations, or who apply incorrect codes with high consistency, can produce high reliability figures without valid analysis (Gwet, 2014: 19, 314).

In humanitarian and development organizations, qualitative coding underpins analysis of needs assessments, community consultations, complaint and feedback mechanisms, and program evaluations. Operational constraints are substantial: multilingual source data, variable translation quality, limited analytical staff, and time pressure routinely limit the depth and coverage of qualitative analysis that organizations can conduct. Earlier efforts to automate coding using machine learning and dictionary-based methods were limited by the need for large domain-specific training datasets and poor generalization across contexts (Nelson et al., 2021; Li et al., 2021). LLMs offer a different capability profile: they can apply a coding framework described in natural language, require no domain-specific labeled training data, and can in principle handle varied topics and communication styles within a single deployment. These properties position LLMs as plausible candidates for humanitarian qualitative coding applications, though whether performance under controlled research conditions transfers to operational deployment remains to be established.

**LLM Reliability for Qualitative Coding**

The clearest pattern across the published literature on LLM-assisted coding is that models perform more reliably on deductive tasks, applying a fixed codebook to new text, than on inductive tasks, where thematic categories must be inferred from the data without prior specification. This contrast has been documented consistently across fields and model families.

In deductive settings, reported agreement can be substantial. GPT-4o achieved mean percent agreement of 96% (SD 4%) and mean Cohen's kappa of $\kappa = 0.71$ (SD 0.26) when applying a human-developed codebook to medical education transcripts; the same study found that when GPT-4 was tasked with inductive code generation, only 31% of produced codes matched human-generated codes, 26% were considered reasonable alternatives, and 42% were judged not reasonable (Jennings et al., 2025).. Comparable deductive performance patterns has been reported in physics education research and learning design (Borse et al., 2025; Rodríguez Triana et al., 2025). Across the literature as a whole, agreement in deductive settings spans $\kappa = 0.46$ to $\kappa = 1.0$, a range wide enough that no individual study's results generalize without qualification (Zhang et al., 2025; Dunivin, 2025).

The contrast between deductive and inductive performance reflects an alignment between LLM architecture and the constrained deductive task. When explicit code definitions limit the classification problem to label assignment within a fixed set, models can apply contextual pattern-matching in a way consistent with their design. When thematic structure must be inferred, models are required to generate interpretive categories from the data rather than apply predefined ones, and the resulting outputs are less consistent. Error audits of deductive coding identify the specific failure modes that arise when contextual inference is required: over-interpretation of tone, failure to track meaning across transcript segments, and confusion between hypothetical and directly experienced events (Jennings et al., 2025). These errors reflect the limits of surface-pattern processing when applied to text that communicates meaning through implication, narrative structure, or contextual reference. In humanitarian data, where participants may express physical safety concerns indirectly or describe discrimination through accounts of service denial, these failure modes carry direct operational significance.

**Workflow Design, Prompting, and Validation**

Reliability is not solely a property of the model; it also depends substantially on the workflow in which the model is deployed. Codebook adaptation, prompt structure, and output validation procedures each materially affect reported performance, such that agreement figures produced under carefully engineered conditions may not transfer to less controlled operational deployment.

Dunivin (2025) argues for rewriting code definitions specifically for machine comprehension, replacing the inferential conventions available to trained human coders with explicit disambiguation criteria, and for prompting models to articulate their coding rationale before assigning a label. This approach produced improved coding fidelity in a reflexive content analysis framework. Structured prompting strategies, including few-shot approaches that pair code definitions with illustrative examples, generally outperform zero-shot instructions in deductive tasks (Rodríguez Triana et al., 2025). Multi-run validation, in which each transcript is coded across multiple independent model runs and the modal response is retained, substantially reduces stochastic variation in outputs (Jain et al., 2025). The choice of reliability coefficient also has methodological relevance: Gwet's AC1 has been proposed as more appropriate than Cohen's kappa under conditions of imbalanced category distributions, where kappa can produce misleadingly low estimates regardless of genuine agreement (Rodríguez Triana et al., 2025).

These workflow features identify actionable levers for improving LLM coding performance in practice. At the same time, they complicate the interpretation of published performance figures. High agreement under carefully engineered evaluation conditions may not generalize to deployment contexts where codebooks are less systematically developed, prompts are less refined, or multi-run validation is not feasible at operational scale. A related concern is the under specification of human benchmarks in many published studies: several evaluations do not report coder training procedures, disagreement resolution methods, or the inter-coder reliability achieved among human coders prior to the LLM comparison, making it difficult to interpret reported agreement between LLM and human coders as a well-defined measure of analytical quality (Belur et al., 2021; Halpin, 2024; Xiao et al., 2023).

**Validity, Accountability, and the Absence of Humanitarian Evidence**

Agreement with a human Gold Standard does not establish that coded outputs are analytically valid. A model may reproduce a human coder's label through a different inferential pathway, such as surface keyword matching, statistical associations from pre-training, or categorical default under uncertainty, and still achieve high agreement while failing to capture the intended meaning of the text across a broader range of conditions. Validity evidence in the existing literature is thinner than reliability evidence. Where error audits are conducted, they identify specific failure modes, including keyword-driven miscoding in which a thematically irrelevant code is applied because a lexically salient term appears in the text (Jennings et al., 2025). Systematic validity testing across diverse corpora and coding conditions is limited or absent in most studies (Xiao et al., 2023; Lin & Zhang, 2025, Zhang et al., 2025). High agreement under controlled conditions is therefore an insufficient basis for deployment decisions where coding outputs inform consequential choices.

Accountability mechanisms are discussed across the literature as aspirational principles rather than operational specifications. Consistent recommendations include human oversight of model outputs, transparent documentation of prompts and codebooks, reproducibility through shared workflows, and expert review of borderline or high-stakes excerpts (Dunivin, 2025; Ornelas et al., 2025; Jain et al., 2025). Some studies implement specific accountability features, including Application Programming Interface (API)-based workflows designed for privacy-sensitive applications (Zhang et al., 2025) and open-source tools with configurable validation parameters (Jain et al., 2025). However, formal governance mechanisms, including audit trails, version-controlled coding outputs, and escalation procedures when model and human coders diverge, are not specified in most studies. The literature endorses human-in-the-loop oversight as essential without defining what adequate oversight requires in practice, particularly when coding is conducted at scale and resources for review are constrained.

Most directly relevant to the present study, no published research reports on LLM-assisted qualitative coding in humanitarian, crisis response, displacement, or public health emergency contexts. The available evidence comes from medical education, physics education research, software engineering qualitative synthesis, and general qualitative methods, which are settings that share structural features with humanitarian data analysis but differ substantially in linguistic complexity, cultural range, thematic sensitivity, and the direct consequences of analytical error (Borse et al., 2025; Jennings et al., 2025; Kreutzer et al. 2025; Lin & Zhang, 2025; Ornelas et al., 2025). Performance in proxy settings does not establish that LLMs can handle the specific challenges of humanitarian qualitative data: multilingual and dialectal variation, communication shaped by trauma or displacement, needs expressed through indirection or cultural idiom, and the costs of miscoding where decisions affect access to assistance. This gap constitutes the primary empirical motivation for the evaluation reported in this paper.

## Methodology

This evaluation study used a three-stage sequential design to assess LLM reliability for qualitative thematic coding of humanitarian data. In the first stage, Krippendorff's alpha was calculated for all 46 models to establish inter-rater reliability relative to a human Gold Standard and assign models to reliability tiers. In the second stage, a discrepancy analysis examined the nature and distribution of coding errors among models in the reliable tier. In the third stage, a qualitative assessment evaluated performance across criteria specific to humanitarian data analysis among the 14 highest-performing models. Together, the three stages provide a statistically grounded, progressively detailed account of model performance that moves from broad reliability classification to fine-grained characterization of error patterns and thematic strengths.

### Study Design and Institutional Context

This evaluation was conducted by KoboToolbox, an open-source humanitarian data collection platform that has developed AI-assisted qualitative coding functionality within its platform.

KoboToolbox therefore has an institutional interest in the findings of this evaluation. The principal safeguard against the risks associated with this position was the separation of roles within the research process: the coders responsible for constructing the human Gold Standard worked independently and were not involved in prompt development for the LLM testing. LLM performance was assessed against the Gold Standard by a lead researcher who reviewed outputs independently before any cross-model comparisons were made. No formal external registration or independent review of the evaluation protocol was undertaken prior to the study. Results are reported for all 46 models without selective omission, including models that performed poorly, and the full results are provided in Appendix 1.

### Synthetic Transcript Corpus

To enable rigorous and repeatable evaluation while protecting the privacy of crisis-affected populations, the evaluation was conducted on a corpus of 150 English-language synthetic transcripts modeled on real-world humanitarian assessments rather than on transcripts from actual interviews with affected individuals. The decision to use synthetic rather than real interview data was governed by two ethical considerations. First, interview transcripts collected in humanitarian field contexts carry consent for purposes of needs assessment and program design, and repurposing such transcripts for AI model evaluation without explicit additional consent from participants would not be appropriate. Second, collecting new transcripts from affected populations specifically for the purpose of AI model testing, with no direct benefit to those populations in the form of humanitarian response, could not be ethically justified. Synthetic data generation resolved both constraints by enabling a realistic and challenging test corpus without drawing on the experiences of vulnerable individuals for purposes that did not serve them directly. Synthetic transcripts are increasingly common in the machine learning and social sciences fields due to privacy concerns (Offenhuber, 2024).

The transcripts were developed through a structured generation process combining researcher input with generative AI, specifically Claude Opus 4.6, across multiple iterations. For each transcript, researchers drafted a scaffold specifying the primary humanitarian need to be expressed, other competing needs, communication style, language complexity, answer relevance, and word count. Researchers also scripted a prompt instructing the LLM to adopt the perspective of a refugee, use idiomatic expressions, include competing needs in some transcripts, and draw on seven publicly available refugee quotes for style and tone. The LLM was instructed to respond to the question: “What needs do you or your household have in this camp that our assistance has not met so far?” (see Appendix 2). Python code was then used to generate transcript text from this scaffold. Generated transcripts were reviewed against actual humanitarian interview transcripts and assessed by researchers with field experience in humanitarian data collection over nine iterations until the corpus was judged to be sufficiently realistic, diverse, and challenging for evaluation purposes. In the end, the human coders reported that the transcripts were not only realistic but challenging to categorize, with some requiring subjective assessments.

Eleven thematic codes were pre-determined: food, medical treatment, physical safety, income or cash, education, water, sanitation and hygiene (WASH), inclusivity, clothing, electricity, heating, or housing. The number of times each theme appeared was generated at random, with the final four codes appearing half as frequently as they were intended to be grouped into “other” during the human and LLM coding stages of the transcript. “Unrelated response” was included as a twelfth code to create transcript content that would not pertain to humanitarian needs. Sixteen different communication styles were randomly assigned, with styles such as “implicit” over-included, to produce more realistic and difficult-to-code transcripts. Answer relevance also varied randomly across four options, from “Completely relevant” to “Not at all relevant,” with “Mostly irrelevant” over-included, since this produced more difficult-to-code transcripts. Finally, language complexity was randomly varied, with “Low” more frequently included than “Medium”; “High” was not included after testing, since that level of complexity generated esoteric narratives (see Appendix 2 for details about transcript generation).

The use of synthetic data introduces constraints on generalizability that require explicit acknowledgment. Performance on synthetic transcripts generated under controlled conditions may not fully represent performance on real humanitarian interview data, which can involve greater linguistic heterogeneity, dialectal variation, translation artifacts, and emotional complexity than synthetic generation reproduces. The findings of this evaluation should therefore be interpreted as establishing a foundation for performance expectations under structured conditions rather than as a direct prediction of operational performance across diverse humanitarian deployment contexts.

The model used to generate the synthetic transcripts, Claude Opus 4.6, was also among the 46 models evaluated in the study, where it ranked among the higher performers. The possibility that transcripts generated by this model may have structural or stylistic features that advantaged it during evaluation cannot be entirely excluded. The research team assessed this risk through comparative review of transcript characteristics and found no systematic patterns that would plausibly confer a classification advantage to Claude Opus 4.6 specifically. This relationship is nonetheless noted as a limitation and should be considered when interpreting results for that model in particular.

### Human Gold Standard Construction

A human Gold Standard for all 150 transcripts was established prior to LLM evaluation. Two coders with experience in qualitative coding of humanitarian interviews worked independently, guided by a shared codebook. Each coder assigned every transcript a primary thematic code from the nine available categories and then rated the relevance of each remaining category to the transcript on an ordinal scale, distinguishing between themes that were relevant, mentioned, or unrelated to the transcript content. The shared codebook contained definitions for each of the nine categories, including instructions for resolving edge cases and conflicts between competing needs.

Where the two coders disagreed on the primary code, a lead researcher reviewed the transcript alongside the coders' independent rationales and adjudicated a final classification. The coders disagreed over 5.3% of the transcripts when identifying the primary need, particularly around the "Inclusivity" and "Other" categories (Krippendorff's alpha = .938). The adjudicated classifications across all 150 transcripts formed the Gold Standard against which all LLM outputs were subsequently compared in Stages 1 and 2.

The codebook used to guide both human coder training and LLM prompt construction specified nine primary categories: Food refers to transcripts in which the primary expressed need concerns access to adequate nutrition, food supplies, variety, or the means to obtain or prepare food. Medical treatment refers to transcripts in which the primary need concerns access to health services, medicine, or care for illness or injury, or outbreaks of illness. Physical safety refers to transcripts in which the primary need concerns protection from violence, conflict physical harm, or theft including violence outside the camp. Income or cash refers to transcripts in which the primary need concerns livelihood, employment, financial resources, or the means to sell goods. Education refers to transcripts in which the primary need concerns access to schooling, learning opportunities, or supplies, and includes religious schooling and higher education. Water, sanitation, and hygiene refers to transcripts in which the primary need concerns access to clean water, sanitation infrastructure including dignified latrines, and hygiene. Inclusivity refers to transcripts in which the primary need concerns discrimination, exclusion from services, unfair treatment, or the feeling of rejection on the basis of identity even if implicit, including cases where discrimination is the barrier to accessing other services such as medical treatment. Other refers to transcripts expressing a primary need that does not correspond to any of the preceding six substantive categories but is relevant. Unrelated response refers to transcripts that do not express a humanitarian need.

**Model Selection**

Forty-six LLMs were purposively selected to represent the range of leading commercially available and open-weights models relevant to humanitarian data analysis as of April 2026. The selection was organized around six providers: Alibaba Cloud, Anthropic, DeepSeek, Google, Mistral AI, and OpenAI. These providers collectively represent the organizations responsible for the most widely deployed general-purpose LLMs as of the evaluation date. Anthropic, Google, and OpenAI constitute the three leading Western commercial AI laboratories and together account for a substantial share of LLM deployment in humanitarian and development sector applications. DeepSeek and Alibaba Cloud represent leading Chinese commercial AI development, with models that have achieved strong performance benchmarks and are increasingly used outside their countries of origin. Mistral AI represents the European open-source focused laboratory landscape and has produced widely used open-weights models relevant to organizations with specific data sovereignty requirements. The inclusion of all six providers was therefore intended to ensure that findings were not specific to a single laboratory tradition or commercial ecosystem.

Twenty unique base model architectures were identified across these six providers, and multiple configurations of each were included where available, covering different reasoning or thinking levels, producing 46 model configurations in total (see Appendix 1 for full list). Open-weights models from five base model architectures, specifically DeepSeek, Gemma, Mistral, Qwen, and GPT-OSS, were intentionally included to enable direct comparison with proprietary models and to assess the viability of open-weights deployment for humanitarian organizations with data governance requirements that limit or preclude transmission of sensitive data to external commercial API services.

### LLM Coding

The authors developed a Python-based web application benchmark testing tool to systematically prompt each LLM to apply available themes using the same codebook wording as was provided to human coders (see Human Gold Standard section above) to evaluate the reliability of LLM-assisted qualitative coding. Researchers defined a benchmark configuration consisting of an interview transcript, a set of analysis questions, expected correct answers, and a selection of LLM models to test. Each question was run asynchronously against the application's LLM pipeline, with results stored in a PostgreSQL database with summary and detailed log data exportable to XLSX.

To mitigate the effect of stochastic variation in model outputs, each LLM coded every transcript seven times under identical prompt conditions. The modal response across the seven runs was used as the model's final code for each transcript, producing 1,050 coding instances per model and 48,300 instances in total. All models were evaluated using a standardized prompt that incorporated the full codebook, including definitions for each category and instructions for resolving conflicts between competing needs. This standardization ensured that performance differences between models reflected differences in the models' internal reasoning rather than differences in instruction quality or prompt framing.

All models were evaluated during a three-day window from April 11 to 13, 2026. The use of a restricted evaluation window was intended to minimize variation in model behavior attributable to provider-side updates or version changes between evaluation runs.

### Stage 1: Krippendorff's Alpha

Inter-rater reliability between each LLM and the human Gold Standard was measured on a nominal scale using Krippendorff's alpha calculated by using a Python script following Marzi et al. (2024). Models achieving $\alpha \geq 0.80$ were classified as reliable, models in the range $\alpha = 0.67$ to 0.79 were classified as tentative, and models with $\alpha < 0.67$ were classified as unreliable, following the inter-coder reliability criteria established by Krippendorff (2019). Krippendorff's alpha compares the observed disagreement in coded data to the expected disagreement under conditions of randomness, producing a coefficient ranging from -1 to 1, with one signifying perfect agreement, zero meaning no agreement beyond randomness, and < 0 conveying systematic disagreement between coders (Krippendorff, 2019; Marzi et al., 2024). This study required the use of only three levels of Krippendorff's alpha (from $\alpha \geq 0.80$ to $\alpha < 0.67$), rather than the full range. This coefficient was selected due to its flexibility; in particular, Krippendorff's alpha handles nearly any sample size and allows for missing data, which was necessary in the limited cases when LLMs produced invalid codes (Krippendorff, 2019; Marzi et al., 2024; Gwet, 2014).

Krippendorff's alpha values and 95% confidence intervals were calculated for all 46 models, with overlapping confidence intervals among reliable-tier models indicating statistical indistinguishability on the reliability measure alone and providing the methodological basis for proceeding to Stage 2 for further differentiation.

**Stage 2: Discrepancy Analysis**

The discrepancy analysis was conducted on all models classified as reliable at Stage 1 to provide a finer-grained account of error types and coding accuracy than Krippendorff's alpha alone. The analysis examined not only whether each LLM code matched the Gold Standard but what kind of disagreement occurred when it did not.

The discrepancy classification drew on ordinal ratings that the human coders had assigned, in addition to primary code assignment, to every remaining thematic category for each transcript. A score of 5 indicated that the category matched the primary code, corresponding to the Gold Standard. Scores of 3 or 4 indicated that the category represented a secondary need that was explicitly framed as important in the transcript. A score of 2 indicated that the theme was present in the transcript but not framed as a need by the speaker while a score of 1 indicated that the theme was not present in the transcript.

Each LLM coding instance was then classified into one of five discrepancy categories on the basis of how the LLM's assigned code corresponded to the human coders' ordinal ratings for that transcript. A Gold Standard response was recorded when the LLM's code matched the human-adjudicated primary code, corresponding to a human rating of 5. A Relevant response was logged when the LLM's code matched a theme the human coders had rated 3 or 4, indicating that the model had identified a secondary need that was prominent in the transcript, but had not correctly prioritized the primary need. A Mentioned response was marked down when the LLM's code matched a theme rated 2, indicating that the model had identified a minor topic present in the transcript but had treated it as a primary need. An Incorrect response was recorded when the LLM's code corresponded to a theme rated 1, indicating a classification with no connection to the transcript content. Finally, an Invalid response was recorded when the LLM failed to return a usable code, for example by producing a blank response or multiple categories.

To produce a single summary metric for cross-model comparison, a Relevance Score was calculated for each model by applying differential weights to the five discrepancy categories. The weighting scheme reflects the diminishing analytical value of each category for humanitarian decision-making. Gold Standard codes received full weight of 1.0, corresponding to exact agreement with the human benchmark. Relevant codes received a weight of 0.5, on the grounds that a model identifying an important secondary need, while not matching the primary classification, still demonstrates thematic comprehension and would direct humanitarian staff toward a valid concern among the population. Mentioned codes were weighted as 0.15, reflecting that a code corresponding to a topic present in the transcript but not a primary need carries minimal operational value for humanitarian organization. Incorrect and Invalid codes received a weight of 0.0, reflecting that entirely erroneous or unusable outputs provide no analytical value and may misdirect humanitarian aid delivery. The Relevance Score for each model was calculated as the weighted proportion of code categories across all 150 transcripts.

**Stage 3: Qualitative Assessment**

The 14 models selected for qualitative assessment were identified from the reliable tier by retaining all base model architectures with $\alpha \geq 0.80$ and selecting only the highest-performing reasoning or thinking configuration based on their Relevance Score. This reduced the 32 reliable models to 14 while preserving representation of the full range of reliable models.

Four criteria were selected to structure the qualitative assessment, each corresponding to a documented challenge in qualitative coding of humanitarian data. The first criterion was the recognition of needs outside the predefined category list, classified under "Other." Humanitarian field data frequently includes accounts of needs that do not map onto standard assessment frameworks, including novel livelihood disruptions, community-specific resource deficits, and needs spanning multiple conventional categories (ACAPS, 2014). A model that defaults to the nearest available category rather than flagging an out-of-scope response as "Other" may affect the analysis and subsequent aid delivery. The second criterion was the ability to recognize discrimination and code inclusivity as the primary need, rather than directly coding the withheld service, such as education or medical treatment. In humanitarian contexts, individuals facing discrimination in accessing services experience a fundamentally different barrier from those lacking services due to resource constraints, and the two require different programmatic responses (Sphere Association, 2018). Miscoding discrimination as a medical or food access need would systematically misrepresent the nature of the barrier and obscure the need for protection-oriented intervention. The third criterion was the ability to process complex needs hierarchies in transcripts where the primary need was expressed subordinately or implicitly in relation to other content. Interview responses in humanitarian contexts frequently embed the most critical need within context or narrative, and models relying on surface salience will misrepresent the speaker's need (Ellsberg and Heise, 2005; Bennett et al., 2016). Finally, the accurate coding of transcripts with slang, idiomatic language, or metaphor was included. As in all communities, crisis-affected populations communicate using culturally-specific language that may not neatly map onto humanitarian assessments–a model processing only literal language will produce systematic errors on data collected across diverse field contexts (Maitland and Xu, 2015; Miner, 2023).

Transcripts containing each of these criteria were then identified to analyze whether the LLMs coded them correctly. Performance was classified as a strength, moderate capability, or weakness of the model based on the proportion of correct codes across the relevant transcripts (see Table 1).

**Table 1: Qualitative Assessment Coding**

| **Criteria** | **No. Relevant Transcripts** | **Thresholds*** | | |
|---|---|---|---|---|
| | | *Strength* | *Moderate* | *Weakness* |
| "Other" needs | 40 | Above 91% | 80–91% | Below 80% |
| Recognize discrimination | 19 | Above 91% | 80–91% | Below 80% |
| Complex hierarchies | 36 | Above 75% | 65–75% | Below 65% |
| Slang, metaphor | 47 | Above 75% | 65–75% | Below 65% |

* The percent of relevant transcripts for which the LLM coded the primary need correctly (not including those where a Relevant need was coded).

Patterns in themes that the LLMs had difficulty recognizing were explored by identifying the categories each model coded incorrectly at disproportionate rates across all 150 transcripts. Finally, outlier transcripts that only one or a few of the 14 models coded correctly or incorrectly were inductively explored to identify explanatory patterns in how specific models handled particular types of content.

## Results

This section reports findings from the three evaluation stages in sequence. First the reliability tiers for all 46 models based on Krippendorff's alpha are presented, followed by the discrepancy analysis for the 14 models selected for focused evaluation. The qualitative assessment of four humanitarian-specific criteria is laid out in the subsequent section. The final section addresses the specific implications of the open-weights finding for humanitarian data governance.

**Reliability Tiers: Krippendorff's Alpha Results**

Of the 46 models evaluated, 32 achieved the reliability threshold of $\alpha \geq 0.80$ and were classified as reliable, six fell within the tentative range of $\alpha = 0.67$ to 0.79, and eight were classified as unreliable with $\alpha < 0.67$. Full results for all 46 models, including all reasoning and thinking configurations tested, are provided in Appendix 1.

Among the 32 reliable models, 95% confidence intervals showed substantial overlap, indicating that differences in Krippendorff's alpha within this tier were not statistically distinguishable for the majority of model pairs. The reliable tier included models from Anthropic, Google, and OpenAI across multiple base model architectures and reasoning configurations, with alpha values ranging from 0.807 to 0.922. The highest alpha value was achieved by Gemini-3.1-pro (reasoning-medium) at $\alpha = 0.922$, and the lowest within the reliable tier by GPT-oss-20b (reasoning-medium) at $\alpha = 0.807$.

The tentative tier included six models spanning four providers. DeepSeek-r1 achieved the strongest performance within this tier at $\alpha = 0.713$. The base configurations of GPT-5.4-mini and Claude-sonnet-4.5, both without reasoning or thinking modes enabled, fell at the lower boundary of the tentative range at $\alpha = 0.672$ and $\alpha = 0.651$ respectively. These base configurations are informative because their thinking-enabled or reasoning-enabled counterparts were classified as reliable, a pattern that is examined below in relation to the effect of reasoning modes on coding performance. Gemma-4-31b without thinking enabled achieved α = 0.733 and also fell in the tentative tier, while its thinking-enabled configuration was classified as reliable.

The unreliable tier comprised eight models, with alpha values ranging from 0.664 to 0.169. Mistral AI models produced the lowest agreement with the Gold Standard in the entire evaluation: Ministral-3-14b achieved α = 0.302 and Mistral-large-3 achieved α = 0.169. Qwen3-235b from Alibaba Cloud achieved α = 0.532, and DeepSeek-v3 achieved α = 0.590. The base configurations of Claude-haiku-4.5 and Gemma-4-26b-a4b fell at the upper boundary of the unreliable tier at α = 0.606 and α = 0.664 respectively. As with the tentative tier, the position of these base model configurations relative to the strong performance of their reasoning-enabled counterparts in the reliable tier is analytically relevant: the same underlying architecture produced substantially different reliability levels depending on whether internal reasoning capability was activated.

In terms of consistency, models in the reliable tier coded each transcript as the same theme across all seven runs. On the other hand, several models in the unreliable tier were inconsistent across runs, with Mistral-large-3 for instance coding one-third of transcripts as two or three different themes.

From the 32 reliable models, 14 were selected for focused analysis. This selection retained all base model architectures with α ≥ 0.80 and included only the top-performing reasoning or thinking configuration for each, reducing the full reliable tier to a set representative of the LLM architectures. These 14 models and their Krippendorff's alpha values are presented in Table 2.

*Table 2: Krippendorff's Alpha and Discrepancy Analysis Results*

| Provider | Model | K's alpha | Gold Standard | Relevant | Mentioned | Incorrect | Invalid | Relevance Score |
|---|---|---|---|---|---|---|---|---|
| Google | **Gemini-3.1-pro** (reasoning-medium) | **0.922** | 92.6% | 6.6% | 0.0% | 0.9% | 0.0% | **95.9%** |
| OpenAI | **GPT-5.4** (reasoning-low) | **0.899** | 90.7% | 9.1% | 0.1% | 0.2% | 0.0% | **95.2%** |
| Google | **Gemini-3-flash** (reasoning-medium) | **0.899** | 90.2% | 8.6% | 0.0% | 1.1% | 0.2% | **94.5%** |
| Anthropic | **Claude-opus-4.6** (thinking-low) | **0.853** | 87.4% | 11.9% | 0.0% | 0.7% | 0.0% | **93.4%** |
| OpenAI | **GPT-oss-120b** (reasoning-medium) | **0.876** | 88.3% | 8.8% | 2.4% | 0.6% | 0.0% | **93.0%** |
| OpenAI | **O3** (reasoning-low) | **0.860** | 87.3% | 10.9% | 1.5% | 0.3% | 0.0% | **93.0%** |
| Google | **Gemma-4-31b** (thinking) | **0.868** | 87.8% | 10.3% | 0.0% | 1.2% | 0.7% | **93.0%** |
| Anthropic | **Claude-sonnet-4.5** | **0.838** | 86.2% | 13.4% | 0.2% | 0.1% | 0.1% | **92.9%** |
| | (thinking-medium) | | | | | | | |
| Google | **Gemma-4-26b-a4b** (thinking) | **0.899** | 89.3% | 7.1% | 0.2% | 1.3% | 2.0% | **92.9%** |
| Anthropic | **Claude-sonnet-4.6** (thinking-medium) | **0.853** | 85.7% | 13.4% | 0.7% | 0.2% | 0.0% | **92.5%** |
| OpenAI | **O4-mini** (reasoning-low) | **0.899** | 87.5% | 7.2% | 2.2% | 1.0% | 2.1% | **91.5%** |
| OpenAI | **GPT-5.4-mini** (reasoning-high) | **0.876** | 86.3% | 9.1% | 1.0% | 3.6% | 0.1% | **91.0%** |
| OpenAI | **GPT-oss-20b** (reasoning-medium) | **0.807** | 83.2% | 11.0% | 3.8% | 1.6% | 0.4% | **89.3%** |
| Anthropic | **Claude-haiku-4.5** (thinking-medium) | **0.815** | 81.6% | 14.9% | 0.7% | 2.4% | 0.5% | **89.2%** |

*Krippendorff's alpha coefficients and Relevance Scores for 14 reliable models selected for focused analysis (all base model architectures α ≥ 0.80, with only the top performing model included); full results in Appendix 1. Gold Standard is the percentage agreement between the LLM and human coders; Relevant is the percentage the LLM coded a theme framed as important but not primary; Mentioned is a theme included but not framed as a need; Incorrect is an irrelevant theme; Invalid is the LLM returning a blank, non-mutually exclusive, or unusable response.*

### Discrepancy Analysis: Accuracy and Error Classification

The discrepancy analysis of the 14 selected models revealed two distinct performance bands within the reliable tier, distinguishable by both Relevance Score and the distribution of error types across the five discrepancy categories.

The first band comprised three models with Relevance Scores between 94% and 96%: Gemini-3.1-pro (reasoning-medium) at 95.9%, GPT-5.4 (reasoning-low) at 95.2%, and Gemini-3-flash (reasoning-medium) at 94.5%. These models achieved Gold Standard agreement on 90.2% to 92.6% of transcripts. When they diverged from the Gold Standard, disagreements fell predominantly in the Relevant category, indicating that errors reflected misidentification of the primary need in favor of a genuinely important secondary need rather than fundamentally erroneous classification. Incorrect rates for these three models ranged from 0.2% to 1.1%. No Invalid responses were produced by Gemini-3.1-pro (reasoning-medium) or GPT-5.4 (reasoning-low). Performance was consistent across the seven repeated runs per transcript, indicating that results were not attributable to stochastic variation.

The second performance band comprised eleven models with Relevance Scores between 89% and 94%. This group included Claude-opus-4.6 (thinking-low) at 93.4%, GPT-oss-120b (reasoning-medium) and O3 (reasoning-low) both at 93.0%, Gemma-4-31b (thinking) at 93.0%, and Claude-sonnet-4.5 (thinking-medium) at 92.9%. Models in this band generally maintained low Incorrect rates, with the exception of GPT-5.4-mini (reasoning-high) at 3.6%, the highest among the 14 selected models. At the lower boundary of this band, GPT-oss-20b (reasoning-medium) and Claude-haiku-4.5 (thinking-medium) achieved Relevance Scores of 89.3% and 89.2% respectively, with Gold Standard agreement rates of 83.2% and 81.6%.

A notable pattern across the 14 models concerns the distribution of Anthropic model outputs relative to models from Google and OpenAI at comparable Relevance Score levels. Claude-opus-4.6 (thinking-low), Claude-sonnet-4.5 (thinking-medium), and Claude-sonnet-4.6 (thinking-medium) consistently produced higher proportions of Relevant codes and lower Gold Standard rates than models from other providers at similar overall Relevance Scores. For example, Claude-sonnet-4.5 (thinking-medium) produced a Gold Standard rate of 86.2% and a Relevant rate of 13.4%, while GPT-oss-120b (reasoning-medium), with a comparable Relevance Score of 93.0%, produced a Gold Standard rate of 88.3% and a Relevant rate of 8.8%. This pattern held consistently across all three Anthropic models in the selection and indicates a systematic tendency to identify a secondary rather than primary need. Because the Relevance Score awards partial credit for Relevant codes, this pattern did not substantially affect overall rankings within the reliable tier. It does, however, indicate a behavioral characteristic that is relevant when selecting a model for applications where precise primary need identification is required.

An Incorrect rate below 1% was applied as a secondary filter to identify models appropriate for deployment contexts in which entirely erroneous codes are operationally unacceptable. The basis for this threshold is that a model producing Incorrect codes at a rate of one per 100 transcripts or above introduces a systematic error source that undermines confidence in the full dataset. A model that assigns a Relevant code, even when not matching the primary need, retains analytical proximity to the transcript content that a reviewing analyst can recognize and adjust. By contrast, an Incorrect code introduces a classification with no basis in the transcript, which risks misdirecting humanitarian analysis if not identified. Applying this filter to the 14 models in Table 1 produces seven models that meet the criterion (see Table 3). Six of these are proprietary models; the seventh, GPT-oss-120b (reasoning-medium), is an open-weights model, a distinction with direct governance implications.

Among the models excluded by this filter, error patterns in the higher-Incorrect models suggest that some failures arose from surface-level keyword association rather than contextual reasoning. In several instances, transcripts mentioning children were coded as education needs even when the narrative content was unrelated to learning, indicating a reliance on lexical salience rather than interpretive assessment of the speaker's primary concern. This pattern was more pronounced in models at the lower end of the reliable tier and in the unreliable tier, where Incorrect rates reached 13.6% for Gemma-4-26b-a4b, 24.7% for Ministral-3-14b, and 44.7% for Mistral-large-3.

The analysis also showed that activating reasoning capabilities consistently yielded Relevance Score improvements of 10 to 15 percentage points when compared to non-reasoning versions of the same base model architecture. The contrast is most clearly illustrated by Claude-sonnet-4.5: the base configuration without thinking achieved a Relevance Score of 79.2% and $\alpha = 0.651$, placing it in the unreliable tier, while the thinking-medium configuration achieved a Relevance Score of 92.9% and $\alpha = 0.838$, placing it firmly in the reliable tier. Analogous patterns were observed for Gemma-4-31b, where the base configuration achieved $\alpha = 0.733$ compared to $\alpha = 0.868$ for the thinking-enabled version, and for Claude-opus-4.6, where the base configuration achieved $\alpha = 0.815$ and a Relevance Score of 91.3%, compared to 93.4% and $\alpha = 0.853$ for the thinking-low configuration. The reasoning mode premium held across architectures and providers, indicating that the internal deliberative step built into reasoning-enabled configurations materially improves thematic prioritization when transcripts present competing needs or require contextual inference. Deploying a base model configuration to reduce cost or processing latency carries a measurable accuracy penalty with direct implications for coding quality in humanitarian applications.

*Table 3: Top-Tier Models with Incorrect Rate Below 1%*

| Provider | Model | Type | K's alpha | Relevance Score |
|---|---|---|---|---|
| Google | Gemini-3.1-pro (reasoning-medium) | proprietary | 0.922 | 95.9% |
| OpenAI | GPT-5.4 (reasoning-low) | proprietary | 0.899 | 95.2% |
| Anthropic | Claude-opus-4.6 (thinking-low) | proprietary | 0.853 | 93.4% |
| OpenAI | GPT-oss-120b (reasoning-medium) | open weights | 0.876 | 93.0% |
| OpenAI | O3 (reasoning-low) | proprietary | 0.860 | 93.0% |
| Anthropic | Claude-sonnet-4.5 (thinking-medium) | proprietary | 0.838 | 92.9% |
| Anthropic | Claude-sonnet-4.6 (thinking-medium) | proprietary | 0.853 | 92.5% |

**Qualitative Assessment: Performance by Humanitarian Criterion**

The qualitative assessment of the 14 models across four criteria revealed variation in thematic performance that is not fully captured by aggregate reliability or Relevance Score metrics. Models with comparable Relevance Scores showed meaningfully different profiles across criteria, indicating that summary metrics are insufficient for model selection in humanitarian applications. Results for all 14 models are presented in Table 4.

Five models demonstrated strength in recognizing discrimination as a primary need under the inclusivity category: Gemini-3.1-pro (reasoning-medium), Gemini-3-flash (reasoning-medium), Gemma-4-31b (thinking), Gemma-4-26b-a4b (thinking), and Claude-opus-4.6 (thinking-low). Claude-sonnet-4.5 (thinking-medium) also demonstrated strength on this criterion despite moderate performance on other dimensions. By contrast, Claude-haiku-4.5 (thinking-medium), GPT-5.4-mini (reasoning-high), O3 (reasoning-low), GPT-oss-120b (reasoning-medium), GPT-oss-20b (reasoning-medium), and O4-mini (reasoning-low) showed weakness, indicating a systematic tendency to recode transcripts involving discrimination under the service category being denied rather than under inclusivity. In humanitarian contexts, this error is consequential because it obscures the protection dimension of the need and would direct programmatic response toward service provision rather than anti-discrimination intervention. Claude-sonnet-4.6 (thinking-medium) and GPT-5.4 (reasoning-low) showed moderate performance. Discrimination recognition was not consistently associated with overall aggregate performance: Claude-opus-4.6 (thinking-low), ranked fourth overall by Relevance Score, demonstrated strength, while GPT-5.4 (reasoning-low), ranked second, showed only moderate performance on this criterion.

No model demonstrated consistent strength at accurately assigning the “Other” code to transcripts expressing needs outside the nine predefined categories. All 14 models were classified as either exhibiting moderate performance or weakness on this criterion. This represents a universal limitation across the evaluated models and providers rather than a failure specific to particular architectures. The practical implication for humanitarian deployment is that real-world interview data will contain needs that do not map cleanly onto any predefined classification scheme, and a systematic tendency to assign the nearest available predefined category rather than flag an out-of-scope response will introduce misrepresentation of the data that analysts may not readily detect without reviewing individual transcripts.

Four models demonstrated strength at identifying the primary need in transcripts where it was expressed subordinately or through contextual implication rather than explicit statement: Gemini-3.1-pro (reasoning-medium), Gemini-3-flash (reasoning-medium), Gemma-4-26b-a4b (thinking), and Claude-opus-4.6 (thinking-low). A second group, including Gemma-4-31b (thinking), GPT-oss-120b (reasoning-medium), O3 (reasoning-low), Claude-sonnet-4.5 (thinking-medium), Claude-sonnet-4.6 (thinking-medium), and GPT-5.4 (reasoning-low), achieved moderate performance. Claude-haiku-4.5 (thinking-medium), GPT-5.4-mini (reasoning-high), O4-mini (reasoning-low), and GPT-oss-20b (reasoning-medium) showed weakness on this criterion. The association between hierarchy processing performance and overall Relevance Score was moderate but not consistent across all models, indicating that contextual prioritization of needs is a partially separable capability from general coding accuracy.

Performance on slang and metaphor handling broadly tracked overall Relevance Score, with the highest-performing models also demonstrating the most robust handling of non-standard communication styles. Gemini-3.1-pro (reasoning-medium), Gemini-3-flash (reasoning-medium), and Claude-opus-4.6 (thinking-low) demonstrated strength. The majority of models achieved moderate performance, indicating adequate but not consistently reliable handling of idiomatic expressions. Claude-haiku-4.5 (thinking-medium), GPT-5.4-mini (reasoning-high), O4-mini (reasoning-low), and GPT-oss-20b (reasoning-medium) showed weakness, consistent with their lower performance across other criteria.

With regards to theme-specific difficulty, physical safety was identified as a source of disproportionate coding errors in 13 of the 14 models, making it the single most consistently difficult category in the evaluation. Income was identified as difficult by 11 models, and water, sanitation, and hygiene by 10. Education appeared among the difficult themes for seven models, predominantly those in the lower performance band, while food was identified as difficult by six models. Medical treatment was identified as difficult by three models. The concentration of abstract, contextually sensitive, or economically embedded needs among the most difficult themes, and the relative ease of concrete, materially specific needs, indicates that variation in theme-specific performance reflects properties of the classification task rather than idiosyncrasies of individual base model architectures. Needs that are frequently expressed through implication, narrative subtext, or culturally specific framing present systematically greater challenges to LLM classification than needs that are expressed through direct and unambiguous statements.

*Table 4: Qualitative Assessment Results*

| Model | Themes recognized with difficulty | Recognized discrimination | Recognized other needs | Accuracy with difficult hierarchies | Accuracy with slang/ metaphors |
|---|---|---|---|---|---|
| Gemini-3.1-pro (reasoning-medium) | Income; safety; WASH* | Strength | Moderate | Strength | Strength |
| Gemini-3-flash (reasoning-medium) | Income; medical; WASH; safety | Strength | Moderate | Strength | Strength |
| Gemma-4-31b (thinking) | Food; income; safety; WASH | Strength | Moderate | Moderate | Moderate |
| Gemma-4-26b-a4b (thinking) | Food; income; safety; WASH | Strength | Moderate | Strength | Moderate |
| Claude-opus-4.6 (thinking-low) | Education; food; income; safety; WASH | Strength | Moderate | Strength | Strength |
| Claude-sonnet-4.5 (thinking-medium) | Education; income; medical; safety; WASH | Strength | Weakness | Moderate | Moderate |
| GPT-5.4 (reasoning-low) | Income; safety; WASH | Strength | Moderate | Moderate | Moderate |
| Claude-sonnet-4.6 (thinking-medium) | Education; food; income; safety; WASH | Moderate | Weakness | Moderate | Moderate |
| Claude-haiku-4.5 (thinking-medium) | Education; food; income; safety; WASH | Weakness | Weakness | Weakness | Weakness |
| GPT-5.4-mini (reasoning-high) | Education; safety; WASH | Weakness | Weakness | Weakness | Weakness |
| O4-mini (reasoning-low) | Income; medical; safety; WASH | Weakness | Moderate | Weakness | Weakness |
| GPT-oss-120b (reasoning-medium) | Income; safety; WASH | Weakness | Moderate | Moderate | Moderate |
| O3 (reasoning-low) | Education; income; safety; WASH | Weakness | Moderate | Moderate | Moderate |
| GPT-oss-20b (reasoning-medium) | Education; food; income; safety | Weakness | Weakness | Weakness | Moderate |

*WASH = Water, sanitation, and hygiene; Results are in absolute terms rather than each model relative to others.

### Open-Weights Models and Proprietary Alternatives

Among the seven models meeting both the reliable tier threshold and the Incorrect rate criterion below 1%, six are proprietary systems operated through commercial API services provided by Anthropic, Google, and OpenAI. The seventh, GPT-oss-120b (reasoning-medium), is an open-weights model released under an open-source license, meaning it can be deployed and operated on independently configured server infrastructure without routing data through the model developer's commercial API.

This distinction has direct implications for humanitarian data governance. Transmitting sensitive interview data to external commercial API services introduces privacy risks that many humanitarian organizations are operationally constrained from accepting, particularly when interview participants have not consented to their data being processed by third-party commercial systems, or where operational security requirements restrict the use of cloud-based data processing infrastructure. Deployment of an open-weights model on a secured, independently managed server eliminates the data transmission risk while preserving model performance. GPT-oss-120b (reasoning-medium) achieved a Relevance Score of 93.0% and $\alpha = 0.876$, performance comparable to several proprietary models that passed the Incorrect rate threshold, and at typically lower per-inference operating cost than API-based alternatives of equivalent capability.

Gemma-4-31b (thinking), also an open-weights model, achieved a Relevance Score of 93.0% and $\alpha = 0.868$, placing it within the second performance band of the reliable tier. Its Incorrect rate of 1.2% marginally exceeded the 1% threshold. The proximity of this figure to the threshold means that with further prompt refinement or context-specific codebook development, Gemma-4-31b (thinking) may represent a viable open-weights alternative in deployment contexts where the Incorrect rate criterion is applied with tolerance for marginal cases. Organizations with strong data sovereignty requirements and constrained API budgets should consider both GPT-oss-120b (reasoning-medium) and Gemma-4-31b (thinking) as primary candidates, with awareness of their respective profiles across the humanitarian-specific criteria documented in Table 4, particularly noting that GPT-oss-120b (reasoning-medium) showed weakness on discrimination recognition, a limitation that would require human oversight to address systematically in any deployment context.

## Discussion

### LLM Performance Relative to Human Coding Standards

The finding that multiple LLMs match average human coders on deductive qualitative coding of humanitarian transcripts is meaningful, but its significance depends on what human coding standard is used as the reference point. The Gold Standard in this evaluation was constructed by two expert coders with specific experience in humanitarian needs assessment, working under controlled conditions with a well-specified codebook and sufficient time for careful deliberation. This benchmark represents more favorable coding conditions than typically exist in operational humanitarian settings, where qualitative data are frequently coded by program staff rather than trained qualitative researchers, under time pressure, without systematic codebook training, and without adjudication mechanisms for resolving disagreements. In this operational context, the performance levels documented here, Gold Standard agreement rates of 87% to 93% among top-performing models, are not merely comparable to expert coding standards: they may exceed the consistency achievable through routine human coding in many humanitarian organizations.

This observation does not argue for the substitution of LLMs for human qualitative judgment. It does, however, reframe the relevant comparison for humanitarian practitioners. Evaluation frameworks that benchmark LLM performance against ideal human coding conditions may systematically understate the practical value of AI-assisted coding in organizations where the realistic alternative is less rigorous, less consistent, and substantially more resource-intensive analysis. The multi-run modal procedure used in this evaluation, in which each transcript was coded seven times with the modal response retained, showed that all the best performing models applied the same code consistently across all iterations, highlighting the fact that top LLMs are both accurate and consistent in their application. This suggests that LLM-assisted coding, when embedded in an appropriate validation workflow, may offer not only scalability but a form of procedural consistency that is difficult to achieve in resource-constrained humanitarian settings.

The evidence is consistent with the broader literature showing that deductive coding is more tractable for LLMs than inductive theme generation (Jennings et al., 2025; Dunivin, 2025). The present evaluation extends this finding to humanitarian data specifically, providing the first direct evidence that the performance levels reported in proxy settings transfer to a domain characterized by greater linguistic diversity, thematic sensitivity, and operational stakes.

**Theme-Specific Performance and the Structure of Humanitarian Communication**

The systematic difficulty LLMs encountered with physical safety, discrimination, and income is not adequately explained as a general model limitation. These three categories share a structural property that distinguishes them from needs that models coded more reliably: they are frequently implied, communicated indirectly, or through culturally embedded framing rather than through explicit declarative statements. A participant describing food insecurity will typically do so in terms that are linguistically unambiguous. A participant experiencing targeted violence, discrimination by service providers, or economic exclusion may communicate those experiences through accounts of what happened to them, through what they did not receive, or through the framing of others' behavior, without naming the underlying need in terms that map onto predefined classification categories. This is not a failure of articulation; it reflects the social conditions under which these needs arise. Physical safety concerns expressed to an interviewer may carry personal risk if disclosed explicitly. Discrimination by service providers may be normalized or attributed to other causes by the speaker. Income precarity may be expressed through its consequences rather than its causes.

LLMs trained on large corpora of predominantly direct, expository text are not well-positioned to recover meaning from this kind of indirect narrative. The keyword-association errors documented in the discrepancy analysis, where transcripts mentioning children were miscoded as education needs, illustrate one form of this limitation: surface lexical patterns are used as proxies for thematic meaning in cases where contextual reasoning would require inference about the speaker's situation and intent. The systematic concentration of errors in the most contextually sensitive humanitarian categories suggests that this limitation is not distributed randomly across themes but is structured by the communicative properties of the data. This has consequences for the use of LLM-assisted coding in humanitarian analysis that extend beyond aggregate accuracy metrics. A dataset coded with high overall accuracy but systematic underperformance on safety and discrimination needs will misrepresent the protection dimension of the humanitarian situation it describes, with direct consequences for the prioritization of protection programming and resource allocation.

## Human-in-the-Loop Requirements: Toward an Operational Specification

The principle that LLM-assisted qualitative coding requires human oversight is endorsed throughout the existing literature but rarely specified in operational terms (Alqazlan et al., 2025; Dunivin, 2025; Ornelas et al., 2025). The present findings provide a basis for moving from the general principle to a more differentiated specification of where and how oversight is required.

The evidence supports a tiered oversight model in which the intensity of human review is calibrated to the theme being coded rather than applied uniformly across an entire dataset. Transcripts coded under categories where top-performing models demonstrated consistent strength, specifically food and medical treatment, may require lighter-touch verification, such as spot-checking a representative sample and reviewing all Invalid or Incorrect-flagged outputs. Transcripts coded under categories where systematic weakness was documented, specifically physical safety, discrimination under the inclusivity category, income, and water, sanitation, and hygiene, warrant systematic human review. Where an LLM codes a transcript under a food need category but the narrative contains subtext relating to safety or discrimination, the model output may suppress rather than surface the most consequential dimension of the speaker's situation.

Operationalizing this tiered approach requires that LLM-assisted coding platforms provide analysts with visibility into not only the assigned primary code but the full discrepancy profile: which secondary themes were rated as Relevant, which as Mentioned, and which transcripts were flagged as producing Incorrect or Invalid outputs. Without this visibility, human oversight is limited to post-hoc verification of individual codes rather than systematic identification of the categories and transcript types where model performance is least dependable. The finding that reasoning-enabled model configurations produced substantially higher performance than base configurations has a direct implication for governance as well as accuracy: organizations deploying base model configurations to reduce operational costs accept a measurable accuracy penalty that increases the scope of human oversight required to maintain analytical quality.

Accountability for coding outputs cannot be located solely in the LLM or solely in the human reviewer. It is distributed across prompt and codebook design, the model configuration selected, the oversight procedure implemented, and the organizational decisions that govern how coded data are used. A governance framework adequate to humanitarian deployment must address each of these components explicitly rather than treating human oversight as a residual safeguard applied to an otherwise automated pipeline.

**Implications for Humanitarian Data Governance**

The data governance implications of this study extend beyond the performance results to questions about the political economy of AI adoption in the humanitarian sector. Six of the seven models meeting both the reliability threshold and the Incorrect rate criterion are operated through commercial API services provided by three Western technology companies. Transmitting interview data collected from crisis-affected populations to these services involves routing sensitive personal information through commercial infrastructure governed primarily by the data protection frameworks of the jurisdictions in which those companies operate, which may not align with the data sovereignty expectations of the communities from which the data were collected, the national regulatory frameworks of the countries in which humanitarian operations are conducted, or the ethical commitments of the organizations responsible for the data.

The position of GPT-oss-120b (reasoning-medium) as the sole open-weights model meeting the full performance criteria is therefore significant for reasons that go beyond cost. Open-weights deployment on independently managed server infrastructure allows humanitarian organizations to maintain full control over data flows, apply their own security configurations, and operate without dependence on commercial API availability or pricing decisions. The performance comparability of GPT-oss-120b (reasoning-medium) with proprietary alternatives at similar Relevance Score levels means that data sovereignty and analytical quality are not in tension in this evaluation: the open-weights option is not a lower-quality compromise but a viable primary recommendation for organizations with data governance requirements that restrict or preclude external API use.

The marginal exclusion of Gemma-4-31b (thinking) by the 1% Incorrect rate threshold suggests that the landscape of high-performing open-weights models is developing rapidly. The evaluation was conducted in April 2026 and model capabilities are advancing at a pace that means specific performance rankings are likely to shift with subsequent model releases. The methodology described in this paper is designed to be repeatable as new models become available, and the governance framework for selecting among open-weights and proprietary alternatives should be treated as a continuing organizational responsibility rather than a one-time evaluation outcome.

LLM coding resulted in a 94% time saving compared to human coders (see Appendix 3). This is operationally significant for humanitarian organizations, but the framing of humanitarian qualitative analysis as a capacity problem requires critical attention. The primary value of qualitative humanitarian data lies in what it communicates about the specific situations and priorities of crisis-affected populations, not in the rate at which it can be categorized. Efficiency gains realized by reducing human engagement with individual transcripts carry the risk of eroding the interpretive contact between analyst and data that gives qualitative analysis its distinctive value. The appropriate frame for LLM-assisted coding is not replacement of human reading and interpretation but reduction of the mechanical burden of initial categorization, releasing analytical capacity for the interpretive and contextual work that automated systems cannot perform.

**Limitations**

Several limitations of this evaluation require acknowledgment. The use of synthetic rather than real interview data, while ethically necessary under the conditions described above, means that performance was assessed on transcripts generated under controlled conditions that may not fully capture the linguistic heterogeneity, translation artifacts, dialectal variation, and emotional register of real humanitarian field data. The extent to which findings generalize to operational deployment on real interview corpora has not been established and represents the most significant outstanding empirical question for future research.

The evaluation was restricted to nine predefined thematic codes applied within a deductive coding framework. Operational humanitarian coding contexts frequently involve larger and more complex codebooks, and the extension of these findings to more expansive classification tasks cannot be assumed. The restriction to a three-day evaluation window in April 2026 means that model performance at other points in time or following provider-side updates may differ from results reported here. The evaluation was conducted in English, and performance on multilingual data, which is characteristic of most humanitarian field operations, was not assessed.

The use of Claude Opus 4.6 to generate the synthetic transcript corpus, combined with its inclusion as one of the 46 models evaluated, introduces a circularity that may have produced a modest performance advantage for that model, as discussed above. While no systematic structural advantage was identified through comparative review, this design constraint should be addressed in future evaluations through the use of a generation model that is not also included in the evaluation set. Finally, the institutional position of KoboToolbox as both the developer of AI-assisted qualitative coding tools and the organization conducting the evaluation is an inherent limitation that the declared safeguards described above can mitigate but not eliminate.

## Conclusion

This study demonstrates that multiple LLMs can perform qualitative thematic coding of humanitarian data at reliability levels comparable to experienced human coders under structured deductive conditions, with top-performing models achieving Krippendorff’s alpha values between 0.853 and 0.922 and Relevance Scores between 92.5% and 95.9%. The findings represent the first direct evidence of LLM coding performance on humanitarian-context data, extending a literature that has until now relied on proxy settings in medical education, physics education, and software engineering.

The three-stage evaluation framework introduced here, combining Krippendorff's alpha reliability tiers with discrepancy analysis using a five-category error classification and qualitative assessment across humanitarian-specific criteria, provides a methodological contribution that extends beyond the specific models tested. Standard inter-coder reliability metrics aggregate performance across all transcripts and categories in ways that obscure consequential variation in error types and thematic accuracy. The discrepancy analysis, and specifically the distinction between Relevant, Mentioned, and Incorrect codes, permits a finer-grained assessment that is operationally relevant: a model that frequently identifies an important secondary need rather than the primary one is meaningfully different from a model that assigns codes with no connection to the transcript content, even when both produce the same aggregate deviation from the Gold Standard.

Three findings have particular significance for humanitarian organizations considering the adoption of LLM-assisted qualitative coding. First, the premium associated with reasoning-enabled model configurations, of approximately 10 to 15 percentage points on the Relevance Score, is large enough to be operationally determinative. Deploying base model configurations to reduce cost or processing time accepts an accuracy penalty that increases the scope of human oversight required to maintain analytical quality and is unlikely to represent a net efficiency gain when oversight costs are factored in. Second, the systematic underperformance of all evaluated models on physical safety, discrimination, and income needs indicates that aggregate accuracy metrics are insufficient for deployment decisions. A model achieving 93% overall Relevance Score may nonetheless misrepresent the protection dimension of a humanitarian dataset in ways that are consequential for programmatic response. Tiered human oversight calibrated to theme-specific performance profiles is therefore necessary rather than optional. Third, the performance comparability of GPT-oss-120b (reasoning-medium) with proprietary alternatives at equivalent reliability levels means that humanitarian organizations with data governance requirements restricting external API use are not required to accept a quality trade-off to maintain data sovereignty. Open-weights deployment on secured infrastructure is a viable and in many respects preferable option for organizations handling sensitive interview data from crisis-affected populations.

The broader significance of these findings is not that LLMs have resolved the scalability problem in humanitarian qualitative analysis. It is that a specific, well-governed application of LLM-assisted deductive coding, one that employs structured prompts, reasoning-enabled configurations, tiered human oversight, and preference for open-weights deployment where data sensitivity requires it, can materially expand the analytical capacity of humanitarian organizations without requiring the epistemological compromises that undifferentiated automation would entail. The conditions under which this is true are specific and demanding. They require investment in codebook development, prompt engineering, and oversight workflows that are not trivial, and they require organizational commitments to data governance that go beyond technical configuration. The evidence supports the feasibility of AI-assisted humanitarian qualitative coding under these conditions. It does not support the conclusion that any capable LLM deployed without these conditions can reliably substitute for human interpretive judgment in contexts where analytical errors may affect decisions about the needs of vulnerable populations.

Future research should address the generalizability of these findings to real interview data, multilingual corpora, and larger classification frameworks. Independent replication by researchers without institutional ties to the tools being evaluated would strengthen confidence in the reported performance levels. Longitudinal evaluation tracking model performance across provider-side updates would address the temporal instability inherent in evaluations of rapidly evolving systems. Most fundamentally, research that follows LLM-assisted coding outputs into organizational decision-making, documenting how coded data are used, where errors propagate, and what governance mechanisms succeed or fail to catch them, is needed before the operational governance frameworks described here can be considered adequate rather than provisional.

## Bibliography

ACAPS (2014) Humanitarian Needs Assessment: The Good Enough Guide, The Assessment Capacities Project (ACAPS), Emergency Capacity Building Project (ECB) and Practical Action Publishing, Rugby, UK. http://dx.doi.org/10.3362/9781780448626

Alqazlan, L., Fang, Z., Castelle, M., & Procter, R. (2025). A novel, human-in-the-loop computational grounded theory framework for big social data. Big Data & Society, 12(2). https://doi.org/10.1177/20539517251347598

Belur, J., Tompson, L., Thornton, A., & Simon, M. (2021). Interrater reliability in systematic review methodology: Exploring variation in coder decision-making. Sociological Methods & Research, 50(2), 837–865. https://doi.org/10.1177/0049124118799372

Bennett, C., Foley, M., & Pantuliano, S. (2016). Time to let go: remaking humanitarian action for the modern era. ODI, London. https://media.odi.org/documents/10422.pdf

Borse, N. S., Subramaniam, R. C., & Rebello, N. S. (2025). Investigation of the Inter-Rater Reliability between Large Language Models and Human Raters in Qualitative Analysis. arXiv preprint https://doi.org/10.48550/arXiv.2508.14764

Braun, V., & Clarke, V. (2006). Using thematic analysis in psychology. Qualitative research in psychology, 3(2), 77-101. https://doi.org/10.1191/1478088706qp063oaDunivin, Z. O. (2025). Scaling hermeneutics: a guide to qualitative coding with LLMs for reflexive content analysis. EPJ Data Science, 14(1). https://doi.org/10.1140/epjds/s13688-025-00548-8 https://link.springer.com/content/pdf/10.1140/epjds/s13688-025-00548-8.pdf

Ellsberg, M. and Heise, L. (2005). Researching violence against women: a practical guide for researchers and activists. World Health Organization and PATH. https://www.who.int/publications/i/item/9241546476

Gwet, K. L. (2014). Handbook of inter-rater reliability: The definitive guide to measuring the extent of agreement among raters. Advanced Analytics, LLC.

Halpin, S. N. (2024). Inter-coder agreement in qualitative coding: Considerations for its use. American Journal of Qualitative Research, 8(3), 23–43. https://doi.org/10.29333/ajqr/14887Jain, N., Hyungil, S., Adeyinka, S., Roseman, L., and Allsop, A. (2025). Multi-LLM thematic analysis with dual reliability metrics: combining Cohen's kappa and semantic similarity for qualitative research validation. arXiv preprint. https://doi.org/10.48550/arxiv.2512.20352 Jennings, K. M., Zahn, A., DeRegnaucourt, A., Taliwal, N., Kelleher, M., Zhou, C. Y., Overla, S., Santen, S. A., Weber, D. E., and Turner, L. (2025). Evaluating LLM assisted qualitative analysis in medical education research: a comparison of human and AI-generated thematic coding (preprint). JMIR Publications Inc. https://doi.org/10.2196/preprints.85572

Kreutzer, T., Orbinski, J., Appel, L., An, A., Marston, J., Boone, E., & Vinck, P. (2025). Ethical implications related to processing of personal data and artificial intelligence in humanitarian crises: a scoping review. BMC Medical Ethics, 26(1), 49. https://doi.org/10.1186/s12910-025-01189-2

Krippendorff, K. (2019). Content analysis: an introduction to its methodology (4th ed.). SAGE.

Li, Z., Dohan, D., & Abramson, C. M. (2021). Qualitative coding in the computational era: A hybrid approach to improve reliability and reduce effort for coding ethnographic interviews. Socius, 7. https://doi.org/10.1177/23780231211062345

Lin, H., & Zhang, Y. (2025). Navigating the Risks of Using Large Language Models for Text Annotation in Social Science Research. Social Science Computer Review, 44(3), 403-427. https://doi.org/10.1177/08944393251366243

Maitland, C., & Xu, Y. (2015). A social informatics analysis of refugee mobile phone use: A case study of Za'atari Syrian refugee camp. TPRC. https://papers.ssrn.com/sol3/Delivery.cfm?abstractid=2588300

Marzi, G., Balzano, M., & Marchiori, D. (2024). K-Alpha calculator–krippendorff's alpha calculator: a user-friendly tool for computing krippendorff's alpha inter-rater reliability coefficient. MethodsX, 12, 102545. https://doi.org/10.1016/j.mex.2023.102545

Miner, A. S., Stewart, S. A., Halley, M. C., Nelson, L. K., & Linos, E. (2023). Formally comparing topic models and human-generated qualitative coding of physician mothers' experiences of workplace discrimination. Big Data & Society, 10(1). https://doi.org/10.1177/20539517221149106

Nelson, L. K., Burk, D., Knudsen, M., & McCall, L. (2021). The future of coding: A comparison of hand-coding and three types of computer-assisted text analysis methods. Sociological Methods & Research, 50(1), 202–237. https://doi.org/10.1177/0049124118769114

Offenhuber, D. (2024). Shapes and frictions of synthetic data. Big Data & Society, 11(2). https://doi.org/10.1177/20539517241249390

Ornelas, T., Araújo, A.A., Araújo, J., Araújo, M., Trinkenreich, B. and Kalinowski, M., 2025. LLM-Assisted Thematic Analysis: Opportunities, Limitations, and Recommendations. arXiv preprint https://doi.org/10.48550/arXiv.2511.14528

Rodríguez Triana, M. J., Saban, M., Asensio-Pérez, J. I., Prieto, L. P., Haba-Ortuo, I., Villa-Torrano, C., & Gillet, D. (2025, September). Code-Aware LLM Prompting in Deductive Qualitative Analysis: A Study in Multi-framework Analysis of Learning Designs. EC-TEL 2025. https://www.researchgate.net/publication/395189459_Code-Aware_LLM_Prompting_in_Deductive_Qualitative_Analysis_A_Study_in_Multi-framework_Analysis_of_Learning_Designs

Sphere Association. The Sphere Handbook: Humanitarian Charter and Minimum Standards in Humanitarian Response, fourth edition, Geneva, Switzerland, 2018. www.spherestandards.org/handbook

Xiao, Z., Yuan, X., Liao, Q. V., Abdelghani, R., and Oudeyer, P.-Y. (2023). Supporting qualitative analysis with large language models: combining codebook with GPT-3 for deductive coding. 28th International Conference on Intelligent User Interfaces, 75–78. https://dl.acm.org/doi/pdf/10.1145/3581754.3584136

Zhang, H., Wu, C., Xie, J., Rubino, F., Graver, S., Cai, J., Kim, C., and Carroll, J. (2025). Exploring inductive and deductive qualitative coding with AI: investigating inter-rater reliability between large language model and human coders. AHFE International, 195. https://doi.org/10.54941/ahfe1006232

# Appendix 1

| Krippendorff's Alpha and Discrepancy Analysis Results across All Models | | | | | | | | |
|---|---|---|---|---|---|---|---|---|
| **Provider** | **Model*** | **Gold Standard** | **Relevant** | **Mentioned** | **Incorrect** | **Invalid** | **Relevance Score** | **K's alpha** |
| Google | **Gemini-3.1-pro** (reasoning-medium) | 92.60% | 6.60% | 0.00% | 0.90% | 0.00% | 95.90% | 0.922 |
| Google | **Gemini-3.1-pro** | 92.40% | 6.30% | 0.00% | 1.30% | 0.00% | 95.50% | 0.914 |
| OpenAI | **GPT-5.4** (reasoning-low) | 90.70% | 9.10% | 0.10% | 0.20% | 0.00% | 95.20% | 0.899 |
| Google | **Gemini-3.1-pro** (reasoning-low) | 91.50% | 6.70% | 0.50% | 1.30% | 0.00% | 94.90% | 0.898 |
| Google | **Gemini-3-flash** (reasoning-medium) | 90.20% | 8.60% | 0.00% | 1.10% | 0.20% | 94.50% | 0.899 |
| OpenAI | **GPT-5.4** (reasoning-medium) | 89.00% | 10.60% | 0.40% | 0.10% | 0.00% | 94.30% | 0.868 |
| Google | **Gemini-3-flash** | 90.20% | 8.10% | 0.00% | 1.30% | 0.40% | 94.20% | 0.89 |
| Anthropic | **Claude-opus-4.6** (thinking-low) | 87.40% | 11.90% | 0.00% | 0.70% | 0.00% | 93.40% | 0.853 |
| OpenAI | **GPT-5.4** (reasoning-high) | 87.00% | 12.00% | 0.70% | 0.40% | 0.00% | 93.10% | 0.852 |
| OpenAI | **GPT-oss-120b** (reasoning-medium) | 88.30% | 8.80% | 2.40% | 0.60% | 0.00% | 93.00% | 0.876 |
| OpenAI | **O3** (reasoning-low) | 87.30% | 10.90% | 1.50% | 0.30% | 0.00% | 93.00% | 0.86 |
| Google | **Gemma-4-31b** (thinking) | 87.80% | 10.30% | 0.00% | 1.20% | 0.70% | 93.00% | 0.868 |
| Anthropic | **Claude-sonnet-4.5** (thinking-medium) | 86.20% | 13.40% | 0.20% | 0.10% | 0.10% | 92.90% | 0.838 |
| Google | **Gemma-4-26b-a4b** (thinking) | 89.30% | 7.10% | 0.20% | 1.30% | 2.00% | 92.90% | 0.899 |
| OpenAI | **O3** | 86.60% | 11.90% | 1.30% | 0.20% | 0.00% | 92.70% | 0.845 |
| OpenAI | **O3** (reasoning-high) | 86.50% | 12.00% | 1.10% | 0.40% | 0.00% | 92.70% | 0.86 |
| OpenAI | **GPT-5.4** (reasoning-xhigh) | 86.40% | 12.20% | 0.70% | 0.70% | 0.10% | 92.60% | 0.844 |
| Anthropic | **Claude-sonnet-4.6** (thinking-medium) | 85.70% | 13.40% | 0.70% | 0.20% | 0.00% | 92.50% | 0.853 |
| OpenAI | **GPT-5.4** | 84.70% | 15.10% | 0.10% | 0.20% | 0.00% | 92.20% | 0.823 |
| OpenAI | **O4-mini** (reasoning-low) | 87.50% | 7.20% | 2.20% | 1.00% | 2.10% | 91.50% | 0.899 |
| OpenAI | **GPT-oss-120b** (reasoning-high) | 85.40% | 11.30% | 2.40% | 0.90% | 0.00% | 91.50% | 0.845 |
| Anthropic | **Claude-opus-4.6** | 83.70% | 15.00% | 0.70% | 0.70% | 0.00% | 91.30% | 0.815 |
| OpenAI | **GPT-5.4-mini** (reasoning-high) | 86.30% | 9.10% | 1.00% | 3.60% | 0.10% | 91.00% | 0.876 |
| OpenAI | **GPT-5.4-mini** (reasoning-xhigh) | 84.60% | 10.20% | 1.00% | 3.90% | 0.40% | 89.80% | 0.852 |
| OpenAI | **GPT-oss-20b** (reasoning-medium) | 83.20% | 11.00% | 3.80% | 1.60% | 0.40% | 89.30% | 0.807 |
| OpenAI | **GPT-oss-120b** (reasoning-low) | 82.60% | 13.00% | 1.40% | 2.80% | 0.30% | 89.30% | 0.807 |
| Google | **Gemini-3-flash** (reasoning-low) | 84.60% | 9.10% | 0.70% | 3.30% | 2.40% | 89.20% | 0.849 |
| Anthropic | **Claude-haiku-4.5** (thinking-medium) | 81.60% | 14.90% | 0.70% | 2.40% | 0.50% | 89.20% | 0.815 |
| Anthropic | **Claude-sonnet-4.6** | 79.60% | 18.40% | 0.70% | 0.70% | 0.70% | 88.90% | 0.775 |
| OpenAI | **GPT-5.4-mini** (reasoning-medium) | 82.50% | 10.50% | 1.00% | 6.10% | 0.00% | 87.90% | 0.815 |
| OpenAI | **O4-mini** | 82.50% | 7.50% | 2.80% | 1.10% | 6.20% | 86.70% | 0.881 |
| OpenAI | **GPT-5.4-mini** (reasoning-low) | 80.70% | 10.40% | 1.30% | 7.60% | 0.00% | 86.10% | 0.786 |
| OpenAI | **O4-mini** (reasoning-high) | 82.00% | 7.10% | 2.50% | 1.10% | 7.20% | 85.90% | 0.888 |
| OpenAI | **GPT-oss-20b** (reasoning-high) | 79.90% | 8.80% | 2.60% | 2.40% | 6.40% | 84.70% | 0.845 |
| DeepSeek | **Deepseek-r1** | 72.30% | 20.60% | 1.00% | 4.90% | 1.30% | 82.70% | 0.713 |
| Google | **Gemma-4-31b** | 76.50% | 9.00% | 2.50% | 11.00% | 1.10% | 81.30% | 0.733 |
| OpenAI | **GPT-5.4-mini** | 69.40% | 21.30% | 1.70% | 6.90% | 0.70% | 80.40% | 0.672 |
| Anthropic | **Claude-sonnet-4.5** | 70.70% | 16.60% | 1.30% | 11.40% | 0.00% | 79.20% | 0.651 |
| OpenAI | **GPT-oss-20b** (reasoning-low) | 70.50% | 14.60% | 2.80% | 9.90% | 2.30% | 78.20% | 0.691 |
| OpenAI | **GPT-4.1** | 68.60% | 18.50% | 0.90% | 12.10% | 0.00% | 77.90% | 0.647 |
| Google | **Gemma-4-26b-a4b** | 70.00% | 14.50% | 1.50% | 13.60% | 0.40% | 77.50% | 0.664 |
| Anthropic | **Claude-haiku-4.5** | 60.70% | 29.30% | 0.70% | 2.00% | 7.30% | 75.40% | 0.606 |
| DeepSeek | **Deepseek-v3** | 58.30% | 21.20% | 1.80% | 8.90% | 9.80% | 69.20% | 0.59 |
| Alibaba | **Qwen3-235b** | 52.00% | 22.70% | 2.90% | 11.60% | 10.90% | 63.80% | 0.532 |
| Mistral AI | **Ministral-3-14b** | 39.20% | 28.60% | 6.70% | 24.70% | 0.90% | 54.50% | 0.302 |
| Mistral AI | **Mistral-large-3** | 29.60% | 20.50% | 3.70% | 44.70% | 1.50% | 40.40% | 0.169 |

Sorted by Relevance Score; Krippendorff's alpha (K's alpha) $\geq 0.80$ reliable; $0.67$ - $0.79$ tentative; $< 0.67$ unreliable.
* Any model without 'thinking' or 'reasoning' indicated refers to the base version without this parameter enabled.

## Appendix 2: Synthetic Transcript Generation

| AI Prompt to Generate Transcripts |
|---|
| For each row, act as a specific refugee respondent orally answering the interview question: "What needs do you or your household have in this camp that our assistance has not met so far?". The outputs should be in the form of a transcript of an audio response. The responses must be realistic and match how people would talk when answering such a question (i.e., direct spoken language, with simple vocabulary, hesitations and conversational fillers, no stylistic flourish, narrative framing or imagery). The response in each row must reflect the word count in the first column and the characteristics in the following columns.<br><br>Handling Themes: Check Theme 2. If Theme 2 is "None", treat Theme 1 as a single, clear priority; focus the entire response on this one need. However, if Theme 2 contains a distinct topic (i.e., is not "None"), you must treat Theme 1 and Theme 2 as valid competing concerns. In this case, do not automatically prioritize Theme 1; instead, allow the narrative flow to determine which need sounds more urgent, even if this creates ambiguity about which is the primary need. Treat Theme 3 as secondary – it should be mentioned briefly or in relation to Themes 1 and 2, but framed as much less important and pressing. It can be mentioned at any point in the transcript (beginning, middle or end).<br><br>Important: the responses should describe how respondents feel (sensory details, emotions, situations, behaviors) without putting a clear, definitive label on it (i.e., don't start the transcript with straightforward responses like "it was the waiting" or "it was the smell"), unless the communication style is direct. If it makes sense for the respondent's profile, you can use loose grammar, sentence fragments, and informal phrasing to mimic a real, unpolished oral response. If the word count ends in 3 or 7, you must include translated idioms that sound slightly off in English to reflect a non-native speaker's direct translation. Otherwise, do not use specific idioms, but strictly maintain the simple vocabulary and spoken hesitations described above.<br><br>Output Format: Output ONLY the transcript text. Do not use quotation marks or introduction labels.<br><br>Use the real transcripts below as inspiration for style, voice, rhythm and tone only (do not copy or paraphrase the specific biographical details or content):<br><br>"I lived in a camp for five years. There was some access to education, but not to higher education. I studied there for one year and then I had to stop, until one of my friends told me about an opportunity to study at a university. My family didn't want me to go ahead with my studies, because in my community it's uncommon for women to get higher education. Yet I didn't let them stop me. Since starting university, my life has changed, and I'm a lot more independent...The situation in the camps surrounding Cox's Bazar is becoming very dangerous. In the past, we had a lot of donations. Now, the Bangladeshi people kidnap Rohingya babies and take money from us. Being a refugee is becoming very risky now, and I |

think it’s going to get worse. Since funding has been cut, informal learning centres in the camps have been closed, so no one can get an education. In the past, I received everything at the camp: food, shelter, basic education, everything. When we came to Bangladesh first, we received a lot of support, but now it’s decreasing, I think because the Bangladeshi people are tired of us...My dream is to keep studying and I’d like to go abroad to do that…We want education but nobody gives us access to higher education. I don’t know if migration will become easier or harder, but I know that people will keep moving, and sometimes they will do so in dangerous ways.”

“It was extremely exhausting getting here for me and my mother..bringing donkeys…days, weeks. At the camp here in Chad, some men showed me less respect simply because I was a woman…May Allah be with us refugees. Along the way, a lot happens, but we have to put our lives above the material.”

“We left our house and ran here for safety. We walked for 14 days before reaching the camp. I am happy here, I can at least sleep peacefully. My children can study. We are not allowed to go outside the camp and find work, money or new clothes. I want to return to Myanmar with full citizenship and the right to move freely.”

“The need is for more food and lots of types of food. My husband is not registered, and so he cannot receive food…or other help. We share the food we receive among us. We have a little garden, but this is not enough. You could do that - you could bring in more food. A lot of people had a stomach bug, diarrhoea at the camp, but we were OK. When we are sick, we can have treatment.”

“I am originally from Bhutan. I’m from the southern part of the Bhutan. The name of the district I was born in is Samchi. Now they changed the name and called it Samtse. So I’m from there. Basically Bhutan is a country based on agriculture so they have subsistence farming, our family had a farm and almost 13 acres of land we used to farm. We grew all kinds of things, corn, rice, millet, all kinds of stuff. We do have a betel nut farm and a lychee farm, which is a nice fruit, and we do have a little tea garden. So many things in the same place. My dad, he had some skills in dairy so we had a community dairy, and he was the secretary of the dairy so he was running the dairy basically. And my grandfather, he was a national assembly member, that means he was a government servant, basically compare it in the U.S. to a senator. So we had many different kinds of things. Let me talk about life in the refugee camp. It was really horrible. The houses were made out of small bamboo and thatch, there was no electricity—the people of America, they don’t imagine a life without an electricity, but here we experience a life without electricity. In Bhutan we had electricity and Bhutan is a really rich country in electricity, it has big hydro projects and sells to India. We had electricity in my house. But here I live a life that is totally dark, very difficult. You could bring electricity, actual sanitation, proper food to eat. Life is very miserable.”

"Somalia. I didn’t do nothin’, just chillin’. I was still child. My family farmers. Livestock. I don’t remember much about Somalia...I went to madrasa. Then war. The most important person in life are my father. He does right thing all time. My parents lives in Kenya. If I speak to my grandchildren, I tell them school and do right thing, keep going."

"Our biggest needs - and they are equal - is food and protection from violence. I was born and raised in the refugee camp. I am teacher; I taught for two years after graduated high school, English...the military came to the camp and they shot fire in the village. They yelled at us not to run, not to flee…We've been eating only rice, and not much. We need vegetables, and the children, they need fruits. We've needed food for a long time."

**Table: Themes among Transcripts (n=150)**

| **Theme 1** | **Transcripts*** | **Theme 2** | **Transcripts*** | **Theme 3** | **Transcripts*** |
|---|---|---|---|---|---|
| Education | 14 | Education | 14 | Education | 9 |
| Food | 15 | Food | 10 | Food | 7 |
| Inclusivity | 14 | Inclusivity | 14 | Inclusivity | 8 |
| Income or cash | 15 | Income or cash | 11 | Income or cash | 7 |
| Medical treatment | 13 | Medical treatment | 18 | Medical treatment | 9 |
| Physical safety | 17 | Physical safety | 12 | Physical safety | 6 |
| WASH | 16 | WASH | 12 | WASH | 6 |
| Unrelated | 14 | Unrelated | 11 | Unrelated | 9 |
| [No theme] | N/A | [No theme] | 23 | [No theme] | 76 |
| Clothing ("Other")** | 8 | Clothing ("Other")** | 5 | Clothing ("Other")** | 1 |
| Electricity ("Other")** | 8 | Electricity ("Other")** | 5 | Electricity ("Other")** | 4 |
| Heating ("Other")** | 6 | Heating ("Other")** | 9 | Heating ("Other")** | 6 |
| Housing ("Other")** | 9 | Housing ("Other")** | 6 | Housing ("Other")** | 2 |

* Number of times the LLM was instructed to produce transcripts with this theme; the number of times the theme was coded by humans or LLMs may differ.
** Combined into "Other" theme for human and LLM coding

| Table: Communication Styles among Transcripts (n=150) | |
|---|---|
| **Communication Style** | **No. transcripts** |
| Abstract | 11 |
| Ambiguous | 6 |
| Concise | 6 |
| Direct | 9 |
| Emotionally expressive | 7 |
| Explicit | 5 |
| Fragmented | 6 |
| Implicit | 17 |
| Indirect | 10 |
| Long-winded | 6 |
| Narrative | 9 |
| Rambling | 8 |
| Sparse | 6 |
| Stream-of-consciousness | 18 |
| Structured | 7 |
| Verbose | 9 |

| Table: Answer Relevance among Transcripts (n=150) | |
|---|---|
| **Answer Relevance** | **No. Transcripts** |
| Completely relevant | 25 |
| Somewhat relevant | 23 |
| Mostly irrelevant | 81 |
| Not at all relevant | 21 |

| Table: Language Complexity among Transcripts (n=150) | |
|---|---|
| **Language complexity** | **No. Transcripts** |
| Low | 96 |
| Medium | 54 |
| High | 0 |

| Table: Lengths among Transcripts (n=150) | |
|---|---|
| **No. Words** | **No. Transcripts** |
| 0-24 | 1 |
| 25-49 | 20 |
| 50-74 | 37 |
| 75-99 | 59 |
| 100-124 | 28 |
| 125-149 | 5 |

## Appendix 3

To calculate time savings, we compared the time it takes KoboToolbox users to generate qualitative analyses in the platform to the time human experts spent coding the transcripts. As background, to request the analysis for one transcript at a time in KoboToolbox takes around 5-10 seconds, depending on how fast the user is at clicking. Meanwhile, coding the 150 transcripts took the human experts an average of 7.5 hours (or 450 minutes) each, including reading the codebook.

To calculate time users spend generating codes for 150 transcripts in KoboToolbox:

$$[150 \text{ (transcripts)} \times 10 \text{ (seconds)}] / 60 = 25 \text{ minutes}$$

To compare the time it would take KoboToolbox users to generate analyses to the time it takes human experts to code the same transcripts:

$$(25\text{-}450)|450| \times 100 = {\sim}94.4\% \text{ decrease}$$

Verifying the generated code for the transcript is considered best practice, whether AI-generated, a human coder checking their own work, or a human coder checking a colleague's work. In the KoboToolbox platform this involves comparing the code to the transcript and clicking a "verify" box. Since verification is required whether AI or human coding, this is not taken into account in the above calculation.